4# Monocular Visual–Inertial SLAM Algorithm Combined with Wheel Speed Anomaly Detection

Peng Gang, Lu Zezao, Chen Shanliang, Chen Bocheng, He Dingxin
Key Laboratory of Image Processing and Intelligent Control, Ministry of Education
School of Artificial Intelligence and Automation, Huazhong University of Science and Technology, Wuhan, 430074, China*Abstract*—To address the weak observability of monocular visual–inertial odometers on ground-based mobile robots, this paper proposes a monocular inertial SLAM algorithm combined with wheel speed anomaly detection. The algorithm uses a wheel speed odometer pre-integration method to add the wheel speed measurement to the least-squares problem in a tightly coupled manner. For abnormal motion situations, such as skidding and abduction, this paper adopts the Mecanum mobile chassis control method, based on torque control. This method use the motion constraint error to estimate the reliability of the wheel speed measurement. At the same time, in order to prevent incorrect chassis speed measurements from negatively influencing robot pose estimation, this paper uses three methods to detect abnormal chassis movement and analyze chassis movement status in real time. When the chassis movement is determined to be abnormal, the wheel odometer pre-integration measurement of the current frame is removed from the state estimation equation, thereby ensuring the accuracy and robustness of the state estimation. Experimental results show that the accuracy and robustness of the method in this paper are better than those of a monocular visual–inertial odometer.

*Index Terms*—wheel speed anomaly detection, motion constraint, pose estimation, visual inertial system, SLAM algorithm## I. INTRODUCTION

THE traditional method for estimating a robot's pose relies on a wheel odometer or inertial measurement unit (IMU) for position estimation [1]. Given the initial pose of an indoor mobile robot, its pose at any later time can be calculated using only the speed information from the wheel odometer. The advantage of this wheeled mileage calculation method is that the positioning accuracy is very high over short times and distances. However, this algorithm relies too much on the accuracy of speed acquisition. Imprecise speed measurement will directly affect pose estimation. Moreover, phenomena such as wheel slipping will occur during actual robot movement, which causes the accumulation of angular errors. Integrating angle information from the inertial sensor can improve the accuracy of angle calculation, which can, to an extent, reduce the angular deviation caused by wheel slips or inaccurate heading angle calculation. However, inertial sensors accumulate drift when running for a long time, so the heading angle obtained from gyro angular velocity integration causes a cumulative time error that is not conducive to the long-term positioning and navigation of the robot. Trajectory estimation algorithms that combine inertial sensors and odometers are better able to estimate a mobile robot's pose when it moves quickly over a short time, but perform poorly in calculating a robot's positioning in the long term.

Cameras, which can accurately estimate poses in static environments or those with low-speed motion, are more stable in the long term than inertial sensors. However, for fast movement over short periods of time, the overlap between two frames of camera data is too small for feature matching, which ultimately affects the accuracy of calculating the robot's positioning. The traditional SLAM algorithm, which uses a camera as its only sensor, fails to cope with low light, weak texture, and fast motion environments. The visual–inertial SLAM algorithm combines a monocular camera and IMU, and has excellent short-term tracking performance, making up for the shortcomings of purely vision-based methods in difficult scenarios, such as those involving fast motion and lighting changes [2]. In the literature [3]–[9], filtering is used to fuse visual and IMU measurements to construct visual–inertial SLAM. These methods use the IMU inertial navigation algorithm for extended kalman filter (EKF) state prediction and visual odometry (VO) to measure updates. A popular filter-based visual inertial odometry (VIO) algorithm is MSCKF [8], which maintains several historical camera poses in the state vector, excluding the position of visual feature points, and uses common multiconstrained updates based on the visual characteristics of relationships. Li *et al.* and Qin *et al.* developed the VINS-Mono open source monocular vision inertial mileage calculation method, which is based on tightly coupled nonlinear optimization [10], [11]. This method

"This work was supported by National Natural Science Foundation of China, No.61672244."
Peng Gang, PhD, Assoc. Prof, Email: penggang@hust.edu.cn; Lu Zezao (Co First Author), Master; Chen Shanliang (Corresponding Author) Master graduate student, Email: m201872734@hust.edu.cn; Chen Bocheng, Master graduate student; He Dingxin, Prof.

combines the IMU pre-integration measurement and visual measurement to obtain the maximum posterior probability estimation problem and uses a nonlinear optimization method to estimate the robot's optimal state.

Although fusing inertial sensor and vision data can be used to compensate for the scale uncertainty and poor fast-motion tracking inherent in purely visual pose estimation methods, the combined method is ineffective in nontextured or weakly lit environments in which the visual sensor cannot obtain useable information. At this time, the visual inertia method is degraded to dead reckoning based on inertial navigation only and pose errors will increase rapidly with time. In [12], the scale observability of monocular vision–inertial odometry on ground-based mobile robots was analyzed in detail. Because the visual inertia method requires acceleration to make the scale measurable, when the robot is purely rotating with constant speed, the lack of acceleration excitation removes the constraint on the scale, resulting in a gradual increase in scale uncertainty and positioning errors. For mobile robots with wheel speed sensors, the camera, inertial sensors, and wheel speed sensors can be fused to solve improve pose estimation robustness in complex scenarios. Guo *et al*. collected data from multiple sets of monocular cameras and odometers, and used the least-squares method to estimate the motion of the camera and odometer in their respective coordinate systems, then obtained the optimal camera-odometer reference offline, before finally using nonlinear optimization to determine the robot's positioning [13]. Heng *et al*. used bundle adjustment (BA) optimization instead of the least-squares method. As BA uses more information, pose estimation accuracy was significantly improved using this technique [14]. Considering factors such as wheel and ground slippage, Kejian *et al*. combined a wheeled odometer, IMU, and vision data in their VINS On Wheels algorithm, which constrains the IMU through the two-dimensional information from the wheeled odometer, thereby improving accuracy and stability; however, because the IMU must initialize the bias and gravity, the effect of this algorithm is not very good on mobile platforms [12].

In order to make full use of the constraints of sensor measurements on pose estimation and improve the accuracy of pose estimation, in [12] and [15], a tightly coupled nonlinear optimization method was used to fuse visual, inertial, and wheel speed sensors for robot pose estimation. These studies demonstrated that when a robot moves with constant acceleration or does not rotate, the scale of the visual inertial odometer and the direction of gravity become unobservable, verifying that the introduction of encoder measurement and soft plane constraints can significantly improve the visual inertial mileage of wheeled robot meter accuracy. However, the experiments did not verify the method's robustness in difficult situations. Wheel slippage and other abnormal conditions that introduce incorrect wheel speed measurements will reduce the system's positioning accuracy.

In [16], a probability-based tightly coupled monocular vision wheel speed SLAM method was proposed. This method uses wheel speed measurements and angular speed measurements from a gyroscope to perform wheel odometer pre-integration in a preprocessing step. Wheel odometer pre-integration, visual measurement, and plane constraint factors are all used to formulate the maximum posterior probability estimation problem and the nonlinear optimal method is used to estimate the robot's position. In addition, this study provided motion tracking strategies for various abnormal sensor states. For example, when a wheel is slipping, only the visual measurement is used for pose estimation; conversely, when visual measurement fails, only the wheel odometer pre-integration measurement is used for motion tracking. These methods have significantly improved the robustness of SLAM systems in complex environments. However, the algorithm in [16] uses only the angular velocity measurements provided by the IMU and does not use acceleration measurements. The lack of acceleration measurement makes it impossible to estimate the absolute attitude of the robot in the ground coordinate system, which makes the algorithm only applicable to horizontal ground. If the robot moves in a sloped environment, the algorithm will not be able to correctly estimate the robot's pose or positioning may fail.

This paper proposes a SLAM algorithm that integrates multiple sensor types: monocular vision sensors, IMUs, and wheel speed sensors. This method uses a tightly coupled method to fuse each sensor's measurement and uses a nonlinear optimization method to maximize the posterior probability to solve the optimal state estimation. It also has loop detection and back-end optimization capabilities.

## II. Wheel Speed Abnormality Detection Based on Torque Control of Mecanum Wheels

The Mecanum wheel chassis has three degrees of freedom, which can be rotated and translated with two degrees of freedom at the same time. It is suitable for narrow and complex environments but has high requirements for ground quality. If the ground is uneven or soft, the Mecanum wheel chassis loses a degree of freedom and seriously affects the dead reckoning of the wheel odometer. This paper adopts a Mecanum wheel moving chassis control algorithm based on torque control. This algorithm can estimate the credibility of the wheel speed measurement using the motion constraint error to detect whether the movement of the Mecanum wheel is abnormal.

### A. Mechanum Wheel Kinematics Model

Figure 1 shows a Mecanum wheel installation structure. The forward direction of the vehicle body is defined as the positive direction of the *x* axis of the chassis, the left of the vehicle body is defined as the positive direction of the *y* axis of the chassis, and the center of the four wheels is defined as the center of the chassis, which is the origin of the wheeled odometer coordinate system. When the chassis advances, the rotation of the wheels is defined as forward rotation.

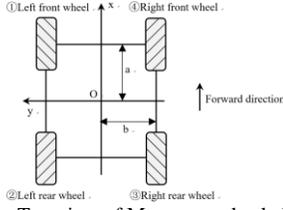

Fig. 1. Top view of Mecanum wheel chassis.

From the three degrees of freedom of the chassis, the kinematics equations of four degrees of freedom of the wheels are solved:

$$\mathbf{v}_{wheel} = \begin{bmatrix} 1 & -1 & -(a+b) \\ 1 & 1 & -(a+b) \\ 1 & -1 & a+b \\ 1 & 1 & a+b \end{bmatrix} \mathbf{v}_{base}, \quad (1)$$

where the expected speed is $\mathbf{v}_{base} = [v_x \ v_y \ \omega]^T$ and the wheel linear speed is $\mathbf{v}_{wheel} = [v_1 \ v_2 \ v_3 \ v_4]^T$.

According to the inverse kinematics equation, if the wheel is rolling without slipping, the wheel speed meets the constraints

$$v_1 + v_4 = v_2 + v_3. \quad (2)$$

When the wheel motion meets the constraint conditions, to ensure robustness, the pseudo-inverse of the inverse kinematics equation is calculated to solve the kinematics equation of the chassis:

$$\mathbf{v}_{base} = \begin{bmatrix} 1 & 1 & 1 & 1 \\ -1 & 1 & -1 & 1 \\ -\frac{1}{a+b} & -\frac{1}{a+b} & \frac{1}{a+b} & \frac{1}{a+b} \end{bmatrix} /4 \cdot \mathbf{v}_{wheel}. \quad (3)$$

If the robot is on uneven ground, one wheel may be off the ground, at which time its speed does not meet the above constraints. In this case, the chassis speed can be calculated from any of the three non-slipping wheels. The following formulas each ignore one of wheels 1, 2, 3, and 4:

$$\mathbf{v}_{base234} = \begin{bmatrix} 0 & 1 & 1 & 0 \\ 0 & 0 & -1 & 1 \\ 0 & -\frac{1}{a+b} & 0 & \frac{1}{a+b} \end{bmatrix} /2 \cdot \mathbf{v}_{wheel} \quad \mathbf{v}_{base134} = \begin{bmatrix} 1 & 0 & 0 & 1 \\ 0 & 0 & -1 & 1 \\ -\frac{1}{a+b} & 0 & \frac{1}{a+b} & 0 \end{bmatrix} /2 \cdot \mathbf{v}_{wheel}$$

$$\mathbf{v}_{base124} = \begin{bmatrix} 1 & 0 & 0 & 1 \\ -1 & 1 & 0 & 0 \\ 0 & -\frac{1}{a+b} & 0 & \frac{1}{a+b} \end{bmatrix} /2 \cdot \mathbf{v}_{wheel} \quad \mathbf{v}_{base123} = \begin{bmatrix} 0 & 1 & 1 & 0 \\ -1 & 1 & 0 & 0 \\ -\frac{1}{a+b} & 0 & \frac{1}{a+b} & 0 \end{bmatrix} /2 \cdot \mathbf{v}_{wheel} \quad (4)$$

*B. Wheel Speed Anomaly Detection*

The method of controlling the speed of the Mecanum wheel chassis is usually based on the inverse kinematics equation, which is used to calculate the desired speed to each wheel, and then independent speed closed-loop control is enacted on each wheel. This speed control method has two disadvantages:

1) Because each wheel adopts independent closed-loop speed control, the output torque of each wheel is uncertain, which may distribute torque unevenly to each wheel. When the torque output of one wheel far exceeds that of the other wheels, the wheel will slip.

2) In the previous section, the Mecanum wheel's motion model explains that the Mecanum wheel chassis can be estimated to be in an abnormal state by checking whether the wheel speed satisfies the movement constraints. If the speed of each wheel is controlled independently, according to the inverse kinematics equation, the observed wheel speed will always meet the motion constraints, which will fail to detect the abnormal motion state of the chassis.

In the following, the covariance of the wheel odometer must be calculated based on the motion constraint error. Therefore, independently controlling the speed of each wheel using to the inverse kinematics equation cannot meet the requirements.

The open differential of robot can evenly distribute power to the left and right wheels, but if one wheel slips, the others will not get enough power. Inspired by the shortcomings of automotive open differentials, this study simulates the effects of automotive differentials through algorithms. The algorithm controls the torque of the chassis by controlling the torque of each wheel independently through a closed loop and controls the speed of the entire chassis. The kinematic equation of a Mecanum wheel based on torque control is as follows:

$$\begin{bmatrix} F_x \\ F_y \\ \tau_o \end{bmatrix} = \begin{bmatrix} \frac{\sqrt{2}}{2} & \frac{\sqrt{2}}{2} & \frac{\sqrt{2}}{2} & \frac{\sqrt{2}}{2} \\ -\frac{\sqrt{2}}{2} & \frac{\sqrt{2}}{2} & -\frac{\sqrt{2}}{2} & \frac{\sqrt{2}}{2} \\ -R & -R & R & R \end{bmatrix} \begin{bmatrix} F_1 \\ F_2 \\ F_3 \\ F_4 \end{bmatrix} = \mathbf{T} \begin{bmatrix} \tau_1 \\ \tau_2 \\ \tau_3 \\ \tau_4 \end{bmatrix}$$

$$\mathbf{T} = \begin{bmatrix} \frac{1}{r} & \frac{1}{r} & \frac{1}{r} & \frac{1}{r} \\ -\frac{1}{r} & \frac{1}{r} & -\frac{1}{r} & \frac{1}{r} \\ -\frac{\sqrt{2}R}{r} & -\frac{\sqrt{2}R}{r} & \frac{\sqrt{2}R}{r} & \frac{\sqrt{2}R}{r} \end{bmatrix} \quad (5)$$

where $R = \sqrt{a^2 + b^2}$ is the radius of the wheel center, $r$ is the wheel radius, and $\mathbf{T}$ is the transformation matrix of the resultant torque of the motor and chassis. Finding the pseudo-inverse of $\mathbf{T}$, we can get the inverse kinematics equation based on torque control:

$$\begin{bmatrix} \tau_1 \\ \tau_2 \\ \tau_3 \\ \tau_4 \end{bmatrix} = \begin{bmatrix} \frac{r}{4} & -\frac{r}{4} & -\frac{\sqrt{2}r}{8R} \\ \frac{r}{4} & \frac{r}{4} & -\frac{\sqrt{2}r}{8R} \\ \frac{r}{4} & -\frac{r}{4} & \frac{\sqrt{2}r}{8R} \\ \frac{r}{4} & \frac{r}{4} & \frac{\sqrt{2}r}{8R} \end{bmatrix} \begin{bmatrix} F_x \\ F_y \\ \tau_o \end{bmatrix}. \quad (6)$$

Using the above formula, the torque of the desired three degrees of freedom of the chassis can be decomposed into the torques of the four motors.

The advantage of torque-based control over speed-based control of the Mecanum wheels is that when a wheel is slipping, its speed is not locked. At this time, the motion constraints are not established, so this method can determine whether the chassis is moving abnormally by checking the motion constraint error. The specific control algorithm is shown in Figure 2.

In Figure 2, at ①, the kinematics solution uses the chassis kinematics equation; at ②, the torque decomposition uses the torque-control-based inverse kinematics equation; at ③, the three proportional–integral controllers are used to perform closed-loop control on the *x*-axis speed, *y*-axis speed, and *z*-axis angular speed, and the controller parameters are obtained by manual parameter adjustment.

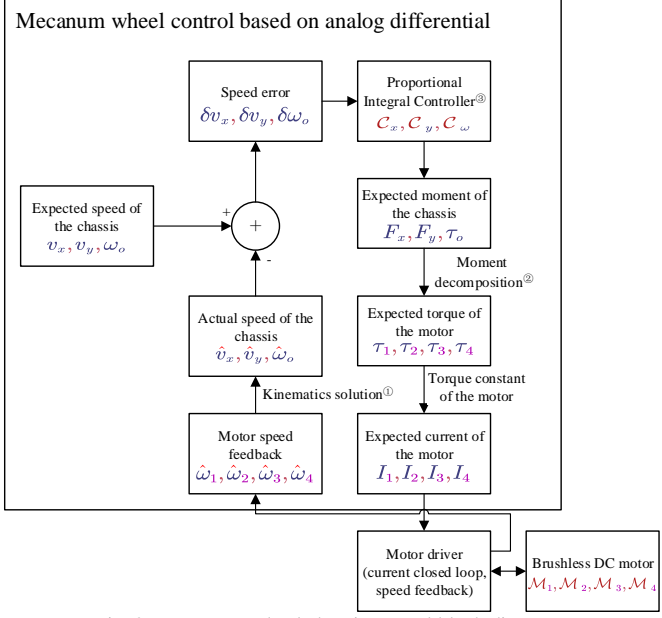

Fig. 2. Mecanum wheel chassis control block diagram.

To verify the performance of the Mecanum wheel chassis speed controller proposed in this paper, the control effect of the controller under step control is experimentally tested. The input and output response curves are shown in Figure 3.

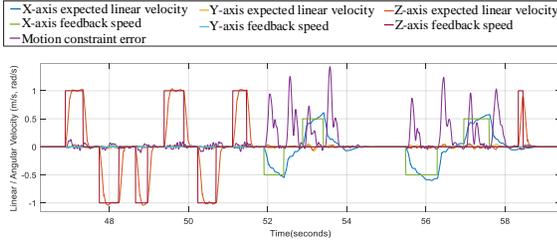

Fig. 3. Response curve of Mecanum wheel controller under step input.

According to the motion constraint error curve in Figure 3 (shown in purple), an obvious motion constraint error appears with step input. In this experimental environment, the error occurs because the wheel output torque exceeds the maximum static friction and causes the wheel to slip. After the wheel slips, the speed of each wheel does not meet the Mecanum wheel movement constraints. Therefore, we cannot determine whether the wheel is slipping by comparing the motion constraint error against a threshold. The Mecanum wheel controller proposed in this paper retains the movement constraint error of the chassis, which can be used to detect abnormal chassis movement.

## III. MONOCULAR INERTIAL SLAM COMBINED WITH WHEEL SPEED ANOMALY DETECTION

### A. Attitude Estimator Combined with Wheel Speed Anomaly Detection

Through the analysis of the kinematics and dynamic characteristics of the Mecanum wheel chassis in the previous section, we found that the reliability of the chassis speed measurement can be estimated by the error of the motion constraint, which can meet the probability-based robot state estimation for the chassis speed measurement demand for variance. The following will analyze the design of a robot state estimator based on the maximum posterior probability combined with wheel speed anomaly detection. The variables to be estimated include the robot's pose, speed, visual feature point depth, and IMU zero bias. Sensor measurements include monocular cameras, an IMU, and mobile chassis speed measurements. First, the maximum posterior probability estimation problem is transformed into a nonlinear least-squares problem. Then, the residual terms corresponding to the constraint factors in the least-squares problem are defined, and the incremental update formula and Jacobian matrix of the IMU pre-integration constraint and the wheel odometer pre-integration constraint are derived. Finally, as ground-based mobile robots usually move on the ground plane, to improve the positioning accuracy when moving on this plane, we define an optional plane constraint factor. To clarify the sequence of each data processing step and the input–output relationship, the data flow of the robot state estimation is shown in Figure 4:

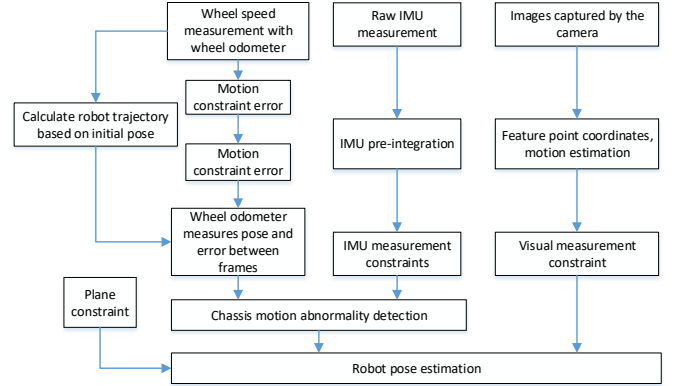

Fig. 4. Robot state estimation structure.

Because the bias of the IMU changes throughout this process, the fixed value obtained by calibration affects the accuracy of the IMU observation. Therefore, the IMU zero offset of each key frame is used as the variable to be estimated for use in the optimization. Variable $\mathcal{X}$ to be estimated is defined as:

$$\begin{aligned}\mathcal{X} &= \{\mathbf{x}_k, \lambda_l\}_{k \in \mathcal{K}, l \in \mathcal{L}} \\ \mathbf{x}_k &= [\mathbf{p}_{B_k}^W, \mathbf{v}_{B_k}^W, \mathbf{q}_{B_k}^W, \mathbf{b}_{ak}, \mathbf{b}_{gk}]\end{aligned}, \quad (7)$$

where $\mathbf{x}_k$ is the IMU state at the $k^{\text{th}}$ key frame; $\mathbf{p}_{B_k}^W$ is the position of the IMU in the world coordinate system; $\mathbf{q}_{B_k}^W$ is the attitude of the IMU coordinate system relative to the world coordinate system (in quaternion form); $\mathbf{v}_{B_k}^W$ is the speed of the IMU in the world coordinate system; $\mathbf{b}_a$ and $\mathbf{b}_g$ in the coordinate system are the zero offset of the accelerometer and of the gyroscope; $\mathcal{K}$ is the key frame in the sliding window; $\mathcal{L}$ is the feature point observed in the key frame; and $\lambda_l$ is the inverse depth (reciprocal of the z-axis coordinate) of feature point $l$ in the camera coordinate system of the key frame that was observed for the first time.

Drawing inspiration from VINS-Mono, we add pre-fusion wheel odometer observations, so observation $\mathcal{Z}$ used to constrain variable $\mathcal{X}$ is defined as:

$$\mathcal{Z} = \{\mathcal{Z}_{C_i}, \mathcal{B}_{ij}, \mathcal{O}_{ij}\}_{(i,j) \in \mathcal{K}}. \quad (8)$$

The visual feature point observation is $\mathcal{Z}_{C_i} = \{\hat{\mathbf{z}}_{il}\}_{l \in \mathcal{L}_i}$, which includes all the feature points $\mathcal{L}_i$ observed in the first key frame; The IMU pre-integration observation is $\mathcal{B}_{ij} = \{\hat{\mathbf{a}}_t, \hat{\boldsymbol{\omega}}_t\}_{t_i \leq t \leq t_j}$, which is obtained by integrating all IMU measurements between the $i^{th}$ key frame and the $j^{th}$ key frame. The pre-fused wheel odometer observation is $\mathcal{O}_{ij} = \{\Delta\hat{\mathbf{m}}_{\text{odom}\,t}, \hat{\mathbf{a}}_{\text{avg}\,t}, \hat{\boldsymbol{\omega}}_{\text{avg}\,t}\}_{t_i \leq t \leq t_j}$, which is obtained from all the pre-fused wheel odometer measurement points between the $i^{th}$ and j-th key frames.

### B. Maximum Posterior Estimation and Least Squares Problem

Given observation $\mathcal{Z}$, the optimal solution of variable $\mathcal{X}$ should satisfy the maximum conditional probability $p(\mathcal{X}|\mathcal{Z})$. Therefore, the optimal estimation problem of $\mathcal{X}$ can be transformed into a maximum a posteriori estimation (MAP) problem:

$$\mathcal{X}^* = \underset{\mathcal{X}}{\operatorname{argmax}}\ p(\mathcal{X}|\mathcal{Z}). \quad (9)$$

In Eq. (9), $\mathcal{X}^*$ represents the optimal solution of variable $\mathcal{X}$. According to Bayes' theorem:

$$p(A \mid B) = \frac{p(A) \times p(B \mid A)}{p(B)}, \quad (10)$$

where $p(A \mid B)$ is the conditional probability of $A$ occurring under a given condition $B$—also known as the posterior probability of $A$, $p(A)$ is the prior probability of $A$—also known as the edge probability, and $p(B)$ is the prior probability of $B$. Bayes' theorem can be summarized as $p(A \mid B) \propto p(B \mid A) p(A)$, where $p(B \mid A)$ is called the likelihood. Therefore, the optimal estimation problem of $\mathcal{X}$ in this paper can be transformed into:

$$\mathcal{X}^* = \underset{\mathcal{X}}{\operatorname{argmax}}\ p(\mathcal{Z}|\mathcal{X}) p(\mathcal{X}), \quad (11)$$

where $p(\mathcal{Z}|\mathcal{X})$ is the conditional probability of occurrence of observation $\mathcal{Z}$ in a given state $\mathcal{X}$, which can be calculated according to the observation equation and observation covariance. $p(\mathcal{X})$ is the prior probability (edge probability) of state $\mathcal{X}$, which here represents the constraint on state $\mathcal{X}$ in the sliding window by historical observations related to historical states that have been removed from the sliding window.

Substituting the definition of observation $\mathcal{X}$ and state $\mathcal{Z}$ into the above formula, we get:

$$\begin{aligned} p(\mathcal{Z}|\mathcal{X}) p(\mathcal{X}) &= p(\mathcal{X}) \prod_{(i,j) \in \mathcal{K}} p(\mathcal{Z}_{C_i}, \mathcal{B}_{ij}, \mathcal{O}_{ij} \mid \mathcal{X}) \\ &= p(\mathcal{X}) \prod_{i \in \mathcal{K}} \prod_{l \in \mathcal{L}} p(\hat{\mathbf{z}}_{il} \mid \mathbf{x}_i, \lambda_l) \prod_{(i,j) \in \mathcal{K}} p(\{\hat{\mathbf{a}}_t, \hat{\boldsymbol{\omega}}_t\}_{t_i \leq t \leq t_j} \mid \mathbf{x}_i, \mathbf{x}_j) \\ &\times \prod_{(i,j) \in \mathcal{K}} p(\{\Delta\hat{\mathbf{m}}_{\text{odom}\,t}, \hat{\mathbf{a}}_{\text{avg}\,t}, \hat{\boldsymbol{\omega}}_{\text{avg}\,t}\}_{t_i \leq t \leq t_j} \mid \mathbf{x}_i, \mathbf{x}_j) \end{aligned} \quad (12)$$

In order to clearly express the relationship between the variables to be optimized and the constraints, the maximum posterior problem is represented by a factor graph; the terms in state $\mathcal{X}$ are used as nodes of the factor graph and the product terms are used as factors of the factor graph, as shown in Figure 5.

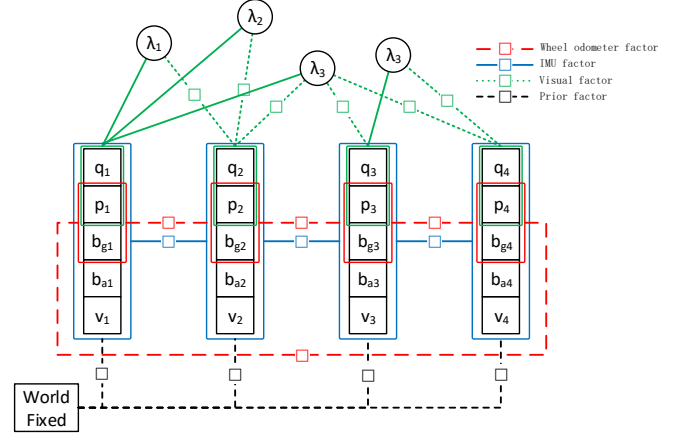

Fig. 5. State estimation factor diagram of robot using vision, inertia, and wheel odometer constraints.

In Figure 5, the circles represent the feature points being tracked, the four boxes in the center represent the state variables of the robot under the four key frames, and the boxes in the lower left corner represent the priors of the states of each key frame in the fixed world coordinate system. Note that the plane constraint is omitted in Figure 5.

The factor graph is optimized by adjusting the value of the nodes so that the product of all factors is maximized. As finding the maximum posterior probability is equivalent to minimizing its negative logarithm, the maximum posterior estimate can be transformed into a least-squares problem:

$$\begin{aligned} \mathcal{X}^* &= \underset{\mathcal{X}}{\operatorname{argmin}}\ -\log[p(\mathcal{X}|\mathcal{Z})] \\ \mathcal{X}^* &= \underset{\mathcal{X}}{\operatorname{argmin}}\ -\log\Bigg[p(\mathcal{X}) \prod_{i \in \mathcal{K}} \prod_{l \in \mathcal{L}} p(\hat{\mathbf{z}}_{il} \mid \mathbf{x}_i, \lambda_l) \prod_{(i,j) \in \mathcal{K}} p(\{\hat{\mathbf{a}}_t, \hat{\boldsymbol{\omega}}_t\}_{t_i \leq t \leq t_j} \mid \mathbf{x}_i, \mathbf{x}_j) \\ &\qquad\qquad \times \prod_{(i,j) \in \mathcal{K}} p(\{\Delta\hat{\mathbf{m}}_{\text{odom}\,t}, \hat{\mathbf{a}}_{\text{avg}\,t}, \hat{\boldsymbol{\omega}}_{\text{avg}\,t}\}_{t_i \leq t \leq t_j} \mid \mathbf{x}_i, \mathbf{x}_j)\Bigg] \\ \mathcal{X}^* &= \underset{\mathcal{X}}{\operatorname{argmin}}\ \Bigg\{-\log[p(\mathcal{X})] - \log\Bigg[\prod_{i \in \mathcal{K}} \prod_{l \in \mathcal{L}} p(\hat{\mathbf{z}}_{il} \mid \mathbf{x}_i, \lambda_l)\Bigg] \\ &\quad -\log\Bigg[\prod_{(i,j) \in \mathcal{K}} p(\{\hat{\mathbf{a}}_t, \hat{\boldsymbol{\omega}}_t\}_{t_i \leq t \leq t_j} \mid \mathbf{x}_i, \mathbf{x}_j)\Bigg] \\ &\quad -\log\Bigg[\prod_{(i,j) \in \mathcal{K}} p(\{\Delta\hat{\mathbf{m}}_{\text{odom}\,t}, \hat{\mathbf{a}}_{\text{avg}\,t}, \hat{\boldsymbol{\omega}}_{\text{avg}\,t}\}_{t_i \leq t \leq t_j} \mid \mathbf{x}_i, \mathbf{x}_j)\Bigg]\Bigg\} \end{aligned} \quad (13)$$

Using the Mahalanobis distance to represent the degree of deviation of the residual from the covariance matrix, we get:

$$\begin{aligned} \mathcal{X}^* = \underset{\mathcal{X}}{\operatorname{argmin}}\ \Bigg\{ & \|\mathbf{r}_p - \mathbf{H}_p \mathcal{X}\|^2 + \sum_{i \in \mathcal{K}} \sum_{l \in \mathcal{L}} \rho\Big(\|\mathbf{r}_C(\hat{\mathbf{z}}_{il}, \mathbf{x}_i, \lambda_l)\|^2_{\Sigma_{C_{il}}}\Big) \\ & + \sum_{(i,j) \in \mathcal{K}} \|\mathbf{r}_B(\{\hat{\mathbf{a}}_t, \hat{\boldsymbol{\omega}}_t\}_{t_i \leq t \leq t_j}, \mathbf{x}_i, \mathbf{x}_j)\|^2_{\Sigma_{\mathcal{B}_{ij}}} \\ & + \sum_{(i,j) \in \mathcal{K}} \rho\Big(\|\mathbf{r}_{\mathcal{O}}(\{\Delta\hat{\mathbf{m}}_{\text{odom}\,t}, \hat{\mathbf{a}}_{\text{avg}\,t}, \hat{\boldsymbol{\omega}}_{\text{avg}\,t}\}_{t_i \leq t \leq t_j} \mid \mathbf{x}_i, \mathbf{x}_j)\|^2_{\Sigma_{\mathcal{O}_{ij}}}\Big)\Bigg\} \end{aligned} \quad (14)$$

where $\|\mathbf{r}\|_{\Sigma}^{2}$ is the Mahalanobis distance of residual $\mathbf{r}$ when the covariance matrix is $\Sigma$. The Mahalanobis distance is defined as:

$$\|\mathbf{r}\|_{\Sigma}^{2} = \mathbf{r}^T \Sigma^{-1} \mathbf{r} . \tag{15}$$

Because visual measurement is easily disturbed by external factors, the Huber loss function $\rho$ can be used to improve the robustness of visual residual $\mathbf{r}_C$ and wheel odometer residual $\mathbf{r}_O$ [17]. The Huber loss function is defined as follows:

$$\rho(r) = \begin{cases} 1, & r \geqslant 1 \\ 2\sqrt{r} - 1, & r < 1 \end{cases} . \tag{16}$$

When the Mahalanobis distance is greater than or equal to 1, or the residual error exceeds 1 standard deviation (the probability of occurrence is less than approximately 32%), the gradient of the residual term for variable $\mathcal{X}$ is 0; that is, there is no longer a constraint on variable $\mathcal{X}$, which prevents outliers from seriously affecting the variables to be estimated and improves the algorithm's robustness.

C. Detection of Chassis Movement Abnormality Combined with the Wheel Speed Sensor

When a ground-based mobile robot experiences motion anomalies such as skidding and abduction, it is possible to estimate the measured covariance through the motion constraint error of the Mecanum wheel chassis. However, when the chassis speed measurement is completely wrong, using these incorrect data in the state estimator will not only fails to improve the positioning accuracy, but also cause the positioning accuracy to decrease, and even cause positioning failure. In order to avoid incorrect chassis speed measurement due to external interference affecting the state estimation effect, this chapter uses three methods to detect the abnormal state of the chassis. When detecting that the chassis is in an abnormal state, the wheel odometer measurement is removed from the state estimation equation to ensure its accuracy and robustness.

The wheeled mileage calculation method assumes that the robot moves on an ideal plane. However, the actual ground may have slopes and undulations, and the two-dimensional wheeled mileage calculation method cannot track motion in three-dimensional space. Introducing the three-dimensional angular velocity measurement provided by the IMU in the wheeled mileage calculation method can not only solve the problem of three-dimensional motion tracking, but also increase the accuracy and reliability of the heading measurement. The wheel speed inertia mileage calculation method uses the wheel speed and angular velocity measurements provided by the IMU for dead reckoning in three dimensions. In this study, the wheel speed inertia mileage calculation method is used between two key frames and the angular velocity measurement of the gyroscope and the position measurement of the wheel odometer are used to calculate the relative pose between the two key frames. This is called the wheeled odometer pre-point. Specifically, the wheel odometer data and the IMU data are first pre-fused, aligned, and packaged into a pre-fused wheel odometer measurement. Then, based on the kinematics equation of the wheel odometer, only the measured values of the pre-fused wheel odometer are continuously calculated and integrated to obtain the relative displacement over time. Finally, the relative displacement obtained by the integration is used as the pre-integral constraint of the wheel odometer, which provides the direction and gradient of variable adjustment for the nonlinear optimization process in the pose estimation of the robot.

The sensor used to measure the wheel speed is unreliable, often leading to very large measurement errors due to uneven ground and wheel slippage. The wheel odometer relies on the wheel speed sensor, and integrates to obtain the relative posture, which is extremely susceptible to the adverse effects of wheel speed measurement errors. In the pre-integration of the wheel odometer, the angle is obtained using the gyroscope to avoid the error caused by the unreliable angle measurement from the wheel odometer. However, displacement errors may still occur due to uneven ground and wheel slippage.

In order to avoid abnormal chassis movement from the source which would adversely affect pose estimation, this paper uses an active detection method to analyze the chassis movement in real time. When the chassis is determined to move abnormally, the wheel odometer pre-integration measurement of the current frame is actively removed from the state estimation equation. In [18], EKF was used to track the wheel's scale factor and, during the data preprocessing process, sensor consensus analysis (SCA) technology was used to measure the wheel speed measurement according to the consistency with the measurement results of other sensors' uncertainty. In order to reduce the false detection rate while guaranteeing a high detection rate of abnormal states, using SCA as in [18], this article also uses three methods to analyze the chassis movement state: Determine whether the chassis is abnormal according to ① the Mecanum wheel movement constraint error; ②the error between the predicted position of the wheel odometer and the predicted position of the IMU state estimator; or ③ the alignment error between the wheel odometer measurement and the IMU measurement. If criterion ① shows abnormal movement or criteria ② and ③ simultaneously indicate abnormal movement, then the chassis is in an abnormal movement state. The three methods of detecting abnormal chassis movements are analyzed below.

1) Detection of Chassis Movement Abnormality Based on Mecanum Wheel Movement Constraints

If the Mecanum wheel chassis moves normally on an ideal plane, the speed of the four wheels should satisfy the motion constraint equation. In real-world scenarios, due to the load of the chassis and the deformation of the rubber roller, the instantaneous rotation speed of the wheel will not satisfy the motion constraint equation and the speed constraint error will change. At this time, if the motion constraint error is integrated over a certain period, the integration result is close to zero.

As the Mecanum wheel chassis moves, if the driving torque of a certain wheel exceeds its maximum rolling friction, it will slip. At this time, the motion constraint error will deviate from the zero position. Therefore, the program in this paper uses the wheel odometer pre-integration algorithm to obtain the cumulative motion constraint error between two frames. If the cumulative motion constraint error exceeds 2 cm and exceeds 1% of the cumulative motion distance, the current chassis is in an abnormal motion state.

Detecting abnormal chassis movement based on the detection of Mecanum wheel movement constraints has disadvantages: when the robot is abducted, when all four wheels are suspended and remain stationary, the movement constraint error is 0. This method cannot detect this abnormal movement state, so the following two detection methods are needed to supplement it.

*2) Detection of Chassis Movement Abnormality Based on Inertial Navigation and Wheel Odometer Consistency*

The multisensor fusion state estimator proposed in this paper can obtain the position, velocity, and attitude of the IMU in the world coordinate system when the camera frame was last received. With the known starting pose and speed, the IMU dead reckoning algorithm, based on inertial navigation, can predict the real-time pose and speed of the carrier in a short time based on the IMU measurement without gravity acceleration.

The wheel odometer pre-integration algorithm can also predict the real-time pose and speed of the chassis based on the previous frame's motion state. According to the position covariance obtained by the wheel odometer pre-integration, the robot's position can be calculated from the IMU or wheel odometer pre-points. If the Markov distance calculated by the two methods is greater than 1.5 (the corresponding probability is approximately 13.4%), the pre-integration result of the wheel odometer is considered abnormal and the chassis is in an abnormal motion state.

To detect abnormal chassis movement, the above method implicitly assumes that the estimated state of the previous frame is accurate. But in many cases, the state estimator cannot give an accurate pose and speed. After the robot performs a pure rotation motion, state estimation accuracy is poor due to the lack of parallax, chassis displacement, and acceleration excitation for the new feature points. If dead reckoning is performed using an inaccurate starting pose and speed, the result of dead reckoning will be greatly offset, causing deviation between the predicted IMU position and the pre-integrated position of the wheel odometer, and normal chassis movement will be misjudged as abnormal.

*3) Detection of Chassis Movement Abnormality Based on Alignment of the Wheel Odometer and IMU*

To avoid the misjudgments caused by the previous abnormal motion detection method without using the state estimation result, this method directly linearly aligns the IMU pre-integration with the wheel odometer pre-integration, and then calculates the deviation between the IMU pre-point and the wheel odometer pre-point in the latest frame. If the deviation of the Mahalanobis distance is greater than 1.5 (the corresponding probability is approximately 13.4%), the pre-integration result of the wheel odometer is considered abnormal and the chassis is in an abnormal motion state.

*D. Plane Constraints*

If it is known that the mobile robot moves on a horizontal plane, the *z*-axis component of the robot's position in the world coordinate system is restricted to 0 in the state estimation problem, which can reduce the degree of freedom of the variables to be estimated and improve the accuracy of the state estimation of the robot. Because there may be small fluctuations on the ground the robot travels, the standard deviation of the *z*-axis component of the robot position is set to 1 cm, based on our experience. The plane constraint residual is defined as:

$$r_\mathcal{P}(\mathbf{x}_k) = \begin{bmatrix} 0 & 0 & 1 \end{bmatrix} \mathbf{p}_{B_k}^W. \quad (17)$$

The Jacobian matrix of residual $r_\mathcal{P}$ with respect to key frame position $\mathbf{p}_{B_k}^W$ is $\mathbf{J}_{\mathbf{p}_{B_k}^W}^{r_\mathcal{P}} = \begin{bmatrix} 0 & 0 & 1 \end{bmatrix}$. Residual $r_\mathcal{P}$ obeys a normal distribution with an expectation of 0 and a standard deviation of 0.01 as $r_\mathcal{P} \sim \mathcal{N}(0, 0.01^2)$. The Mahalanobis distance corresponding to residual $r_\mathcal{P}$ is $\sqrt{\|r_\mathcal{P}\|_{0.01^2}^2} = \sqrt{r_\mathcal{P}^2 / 0.01^2}$.

## IV. EXPERIMENTAL PROCESS AND ANALYSIS

In order to verify the robustness of the SLAM algorithm proposed in this paper, we tested it under a variety of abnormal situations, such as sensor measurement errors or even loss. Because there are overlaps in the types of anomalies that occur in multiple anomalies, the experiment for each anomaly was repeated only once.

*A. Experiments Based on Wheel Slippage*

In the process of controlling the robot's movement in the laboratory, special control instructions are sent to drive the chassis to repeatedly advance and retreat with great acceleration, to force the wheels to slip.

In Figure 6, after the robot normally moves from the origin to the right of the origin, it starts to perform a forced slip operation to make the robot move forward and backward repeatedly, so there is a repeated motion trajectory on the right side of the path diagram. The blue motion path in Figure 6 was obtained by observing the wheeled mileage calculation method. The blue curve shows that the heading angle error increases rapidly after the slip occurs, which causes the subsequent trajectory to be seriously incorrect. The red path in Figure 6 was obtained by observing the wheel speed inertial mileage calculation method. The red curve shows that the heading angle error has basically not increased after the slip occurrence, but the position error has increased. This result shows that the introduction of IMU angular velocity measurement can significantly improve the dead reckoning accuracy of the wheeled mileage calculation method. Comparing the pose errors of the algorithms in Table 2, the VINS-Mono algorithm does not use the wheel speed measurement, so it will not be affected by wheel slippage, and its pose error is the smallest. The accuracy of the multisensor fusion pose estimator proposed in this paper is equivalent to that of VINS-Mono, which indicates that the wrong wheel speed measurement was successfully isolated during the slip process, and the proposed algorithm for detecting abnormal chassis movement is effective.

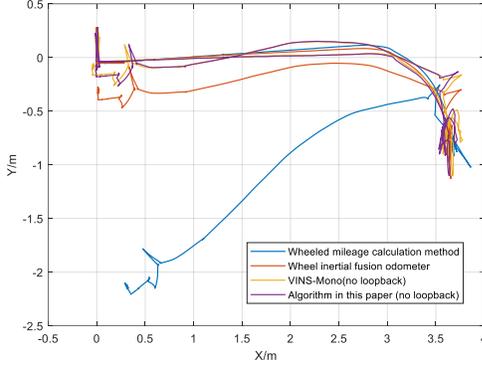

Fig. 6. Path when the wheel is slipping (the robot starts from the origin in the positive direction of the *x* axis).

TABLE I
CHASSIS MOTION PARAMETERS

| Parameter | Value |
|---|---|
| Abnormality duration | 0.100 s |
| Run time | 97.1 s |
| Average speed | 0.071 m/s |
| Maximum speed | 0.760 m/s |
| Displacement | 15.149 m/s |
| Angle | 1653.913 ° |

### B. Robot Collision Experiment

In the process of controlling the robot's movement in the laboratory, special control instructions are sent to drive the chassis to slowly approach the corner of the table. After a collision, some wheels are forced to slip by continuing to drive the chassis forward.

In Figure 7, the robot collided at coordinates (2.2, -0.2). After the collision, the wheels are still rolling forward, so the path obtained by the fusion of the wheel odometer and wheel IMU has been incorrectly extended for a distance after the collision. Figure 7 shows that the VINS-Mono algorithm is not affected by the collision. The algorithm in this paper is also not affected by the collision because it isolates the incorrect chassis speed measurement.

TABLE II
POSTURE ESTIMATION RESULTS WHEN THE WHEELS ARE SLIPPING

| Location Algorithm | Position error | Position error rate | Heading angle error |
|---|---|---|---|
| Wheel Odometer | 2.137 m | 14.11% | 32.497 ° |
| Wheel speed inertial odometer | 0.293 m | 1.94% | 1.792 ° |
| VINS-Mono (no loopback) | 0.078 m | 0.52% | 0.331 ° |
| VINS-Mono (loopback optimization) | 0.005 m | 0.03% | -0.289 ° |
| Algorithm in this paper (no loopback) | 0.085 m | 0.56% | 0.604 ° |
| Algorithm in this paper (loopback optimization) | 0.007 m | 0.05% | 0.065 ° |

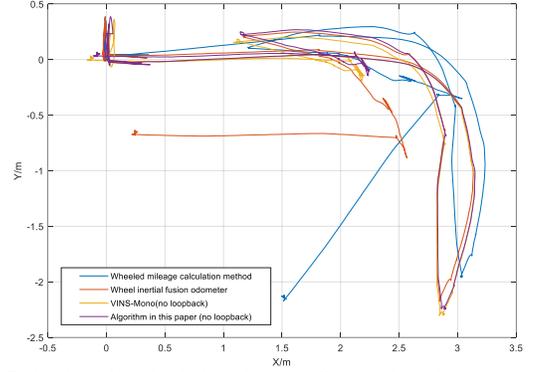

Fig. 7. Path when the wheel slips due to collision (the robot starts from the origin in the positive direction of the X axis).

This experiment verifies that the SLAM algorithm in this paper can still accurately estimate the robot's pose in the case of severe wheel slippage caused by chassis collision.

TABLE III
CHASSIS MOTION PARAMETERS DURING THE EXPERIMENT

| Parameter | Value |
|---|---|
| Abnormal duration | 0.999 s |
| Run time | 111.2 s |
| Average speed | 0.068 m/s |
| Maximum speed | 0.715 m/s |
| Displacement | 17.064 m/s |
| Angle | 1428.642 ° |

TABLE IV
POSTURE ESTIMATION RESULTS WHEN WHEELS COLLIDE DUE TO COLLISION

| Location Algorithm | Position error | Position error rate | Heading angle error |
|---|---|---|---|
| Wheel Odometer | 2.617 m | 15.34% | 56.209° |
| Wheel speed inertial odometer | 0.705 m | 4.13% | 0.231° |
| VINS-Mono (no loopback) | 0.121 m | 0.71% | 0.398° |
| VINS-Mono (loopback optimization) | 0.010 m | 0.06% | 0.436° |
| Algorithm in this paper (no loopback) | 0.082 m | 0.48% | 0.691° |
| Algorithm in this paper (loopback optimization) | 0.022 m | 0.13% | 0.629° |

### C. Wheel Abduction Experiment

In this experiment, the dataset was modified to simulate the abduction of a robot wheel. First, in a complex laboratory environment, the robot is controlled to walk through all channels at a speed of approximately 0.5 m/s and the dataset obtained in the experiment is recorded. Then, the dataset recorded in the experiment was modified to make the wheel odometer speed output 0 after approximately 10 s and the robot pose output is unchanged. The subsequent pose output is also modified to ensure pose continuity.

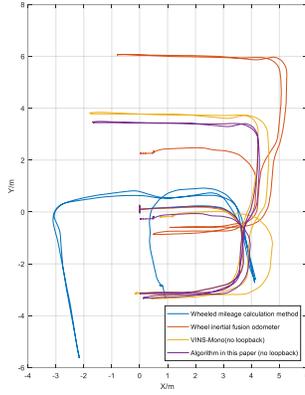

Fig. 8. Path when the robot is abducted (the robot departs from the origin in the positive direction of the *x* axis).

In Figure 8, the robot abduction starts at coordinates (3.8, -0.4) and ends at coordinates (3.8, -2.4). During the kidnapping of the robot, the wheel speed was measured as 0, so the position measurement of the wheeled odometer and the wheel inertial fusion odometer did not change, which is inconsistent with the actual situation, leading to serious errors in subsequent paths.

TABLE V
CHASSIS MOTION PARAMETERS DURING THE EXPERIMENT

| Parameter | Value |
| --- | --- |
| Abnormal duration | 8.391 s |
| Run time | 184.6 s |
| Average speed | 0.266 m/s |
| Maximum speed | 0.613 m/s |
| Displacement | 51.542 m/s |
| Angle | 3428.317 ° |

TABLE VI
POSTURE ESTIMATION RESULTS WHEN WHEELS COLLIDE DUE TO COLLISION

| Location Algorithm | Position error | Position error rate | Heading angle error |
| --- | --- | --- | --- |
| Wheel Odometer | 3.414 m | 6.62% | 101.718° |
| Wheel speed inertial odometer | 2.258 m | 4.38% | -0.507° |
| VINS-Mono (no loopback) | 0.778 m | 1.51% | -0.314° |
| VINS-Mono (loopback optimization) | 0.006 m | 0.01% | -0.725° |
| Algorithm in this paper (no loopback) | 0.275 m | 0.53% | -0.215° |
| Algorithm in this paper (loopback optimization) | 0.010 m | 0.02% | -0.433° |

In wheel slip experiments, robot collision tests, and wheel abduction experiments, due to wheel slippage and missing data, the wheeled mileage calculation method and wheel speed inertia mileage calculation method have large positioning errors. However, the pose estimation of the SLAM algorithm in this paper is not affected. This shows that the multisensor fusion pose estimation algorithm proposed in this paper can accurately identify abnormal chassis movement in a variety of situations and remove the affected wheel odometer pre-integration observation from the pose estimation equation, thereby achieving higher positioning precision. In this paper, the SLAM algorithm can extract valid parts from unreliable wheel odometer measurements to improve positioning accuracy and robustness.

## V. CONCLUSION

This paper reports the design of a monocular inertial SLAM algorithm that functions by detecting abnormal wheel speeds in a mobile robot. Aiming at the disadvantages of traditional Mecanum wheel chassis control algorithm, which eliminates the motion constraint error, this paper designs and implements a Mecanum wheel chassis control algorithm based on a simulated automobile differential. The reliability of the chassis speed measurement can be estimated by calculating the motion constraint error, which can assist in judging whether the chassis is slipping. In order to prevent inaccurate chassis speed measurements from adversely affecting the quality of position estimation, we have used three methods to detect the movement status of the robot's chassis, which can effectively isolate the negative impact of incorrect odometer measurement data on robot state estimation. To verify our algorithm and observe its effects, we conducted wheel slip, robot collision, and wheel abduction tests. The results of these experiments show that the algorithm proposed in this paper can accurately identify abnormal chassis movement in a variety of situations and remove the affected wheel odometry pre-integration observation from the pose estimation equation, thereby improving the accuracy and robustness of mobile robot positioning.

REFERENCES

[1] B. Barshan and H. F. Durrant-Whyte, "Inertial navigation systems for mobile robots," *IEEE Trans. Robot. Autom.*, vol. 11, no. 3, pp. 328–342, 1995.
[2] D. Scaramuzza and F. Fraundorfer, "Visual odometry," *IEEE Robot. Autom. Mag.*, vol. 18, no. 4, pp. 80–92, 2011.
[3] S. Weiss, M. W. Achtelik, S. Lynen, M. Chli, and R. Siegwart, "Real-time onboard visual-inertial state estimation and self-calibration of MAVs in unknown environments," in *Proc. 2012 IEEE Int. Conf. Robotics and Automation*, St. Paul, MN, pp. 957–964.
[4] S. Lynen, M. W. Achtelik, S. Weiss S, M. Chli, and R. Siegwart, "A robust and modular multi-sensor fusion approach applied to MAV navigation," in *2013 IEEE/RSJ Int. Conf. Intelligent Robots and Systems* Tokyo, pp. 3923–3929.
[5] P. Piniés, T. Lupton, S. Sukkarieh, and J. D. Tardos, "Inertial aiding of inverse depth SLAM using a monocular camera," in *Proc. 2007 IEEE Int. Conf. Robotics and Automation*, Rome, pp. 2797–2802.
[6] M. Kleinert and S. Schleith, "Inertial aided monocular SLAM for GPS-denied navigation," in *2010 IEEE Conf. Multisensor Fusion and Integration*, Salt Lake City, UT, pp. 20–25.
[7] E. S. Jones and S. Soatto, "Visual-inertial navigation, mapping and localization: A scalable real-time causal approach," *Int. J. Robot. Res.*, vol. 30, no. 4, pp. 407–430, 2011.
[8] A. I. Mourikis and S. I. Roumeliotis, "A multi-state constraint Kalman filter for vision-aided inertial navigation," in *Proc. 2007 IEEE Int. Conf. Robotics and Automation*, Rome, pp. 3565-3572.
[9] M. Li and A. I. Mourikis, "High-precision, consistent EKF-based visual-inertial odometry," *Int. J. Robot. Res.*, vol. 32, no. 6, pp. 690–711, 2013.
[10] P. Li, T. Qin, B. Hu, F. Zhu, and S. Shen, "Monocular visual-inertial state estimation for mobile augmented reality," in *2017 IEEE Int. Symp. Mixed and Augmented Reality*, Nantes, pp. 11–21.
[11] T. Qin, P. Li, and S. Shen, "VINS-Mono: A robust and versatile monocular visual-inertial state estimator," *IEEE Trans. Robot.*, vol. 34, no. 4, pp. 1004–1020, 2018.

[12] K. J. Wu, C. X. Guo, G. Georgiou, and S. T. Roumeliotis, "VINS on wheels," in *IEEE Int. Conf. Robotics and Automation*, Singapore, pp. 5155–5162.
[13] T. B. Karamat, R. G. Lins, S. N. Givigi, and A. Noureldin, "Novel EKF-based vision/inertial system integration for improved navigation," *IEEE Trans. Instrum. Meas.*, vol. 67, no. 1, pp. 116–125, 2018.
[14] J. Qian, B. Zi, D. Wang, Y. Ma, and D. Zhang, "The design and development of an omni-directional mobile robot oriented to an intelligent manufacturing system," *Sensors*, vol. 17, no. 9, p. 2073, 2017.
[15] F. Zheng, H. Tang, and Y. H. Liu, "Odometry-vision-based ground vehicle motion estimation with SE(2)-constrained SE(3) poses," *IEEE Trans. Cybern.*, vol. 49, no. 7, pp. 2652–2663, 2018.
[16] M. Quan, S. Piao, M. Tan, and S.-S. Huang, "Tightly-coupled monocular visual-odometric SLAM using wheels and a MEMS gyroscope," arXiv preprint arXiv:1804.04854, 2018.
[17] P. J. Huber, *Robust Estimation of a Location Parameter*, New York, NY, USA: Springer, 1992.
[18] A. W. Palmer and N. Nourani-Vatani, "Robust odometry using sensor consensus analysis," in *2018 IEEE/RSJ Int. Conf. Intelligent Robots and Systems*, Madrid, pp. 3167–3173.

**Peng Gang** (1973-) received the Ph.D. degree in engineering from Huazhong University of Science and Technology(HUST), Wuhan, China in 2002.Currently, he is an associate professor in the Department of Automatic Control, School of Artificial Intelligence and Automation, Huazhong University of Science and Technology. A backbone teacher, a member of the Intelligent Robot Professional Committee of the Chinese Artificial Intelligence Society, a member of the China Embedded System Industry Alliance and the China Software Industry Embedded System Association, a senior member of the Chinese Electronics Association, and a member of the Embedded Expert Committee.

**Lu Zezao** (1994-) received B Eng. degree in automation from Center South University, Changsha, China, in 2016. He received master degree at the Department of Automatic Control, School of Artificial Intelligence and Automation (AIA), HUST. His research interests are intelligent robots and perception algorithms.

**Chen Shanliang** (1993-) received his B Eng. degree in automation from Wuhan University of Science and Technology, Wuhan, China, in 2018. He is currently a graduate student at the Department of Automatic Control, School of AIA, HUST. His research interests are intelligent robots and perception algorithms.

**Chen Bocheng** (1996-) received B Eng. degree in automation from Chongqing University, Chongqing, China, in 2018. He is currently a graduate student at the Department of Automatic Control, School of AIA, HUST. His research interests are intelligent robots and perception algorithms.